\def\year{2021}\relax
\documentclass[letterpaper]{article} 
\usepackage{aaai21}  
\usepackage{times}  
\usepackage{helvet} 
\usepackage{courier}  
\usepackage[hyphens]{url}  
\usepackage{graphicx} 
\usepackage{natbib}  
\usepackage{caption} 
\usepackage{cite}
\usepackage{amsmath,amssymb,amsfonts}
\usepackage{algorithmic}
\usepackage{textcomp}
\usepackage{xcolor}
\usepackage{booktabs}
\usepackage{array}
\usepackage{multirow}
\usepackage{multicol}
\usepackage{enumitem}
\usepackage{subfig}

\newcommand{\QSet}{\mathcal{Q}}   
\newcommand{\ASet}{\mathcal{A}} 
\newcommand{\USet}{\mathcal{U}}  

\newcommand{\ZSet}{\mathbb{Z}}
\newcommand{\VoteSet}{\mathcal{V}} 
\newcommand{\Vocab}{\mathcal{W}}

\newcommand{\Loss}{\mathcal{L}}

\newcommand{\eVEC}{\mathbf{e}}
\newcommand{\bVEC}{\mathbf{b}}
\newcommand{\zVEC}{\mathbf{z}}

\newcommand{\AMAT}{\mathbf{A}}
\newcommand{\EMAT}{\mathbf{E}}
\newcommand{\HMAT}{\mathbf{H}}
\newcommand{\WMAT}{\mathbf{W}}

\newcommand{\FC}{\mathrm{FC}}

\newcommand{\ReLU}{\mathrm{ReLU}}

\newcommand{\Softmax}{\mathrm{Softmax}}
\newcommand{\MRR}{\mathrm{MRR}}

\newcommand{\DCG}{\mathrm{DCG}}
\newcommand{\IDCG}{\mathrm{IDCG}}

\newcommand{\Argmax}{\mathrm{argmax}}

\newcommand{\rel}{\mathrm{rel}}

\title{Graph-Based Tri-Attention Network for Answer Ranking in CQA}
\author{
    Wei Zhang\textsuperscript{\rm 1}, Zeyuan Chen\textsuperscript{\rm 1}, Chao Dong\textsuperscript{\rm 1}, Wen Wang\textsuperscript{\rm 1}, Hongyuan Zha\textsuperscript{\rm 2}, Jianyong Wang\textsuperscript{\rm 3}\\
}
\affiliations{
    \textsuperscript{\rm 1}School of Computer Science and Technology, Shanghai Institute for AI Education, East China Normal University\\

    \textsuperscript{\rm 2}School of Data Science, Shenzhen Research Institute of Big Data, The Chinese University of Hong Kong, Shenzhen\\
    \textsuperscript{\rm 3}Department of Computer Science and Technology, Tsinghua University\\
    zhangwei.thu2011@gmail.com, chenzyfm@outlook.com, \{51174500084, 51164500120\}@stu.ecnu.edu.cn 

}

\begin{document}

\maketitle

\begin{abstract}
In community-based question answering (CQA) platforms, automatic answer ranking for a given question is critical for finding potentially popular answers in early times.
The mainstream approaches learn to generate answer ranking scores based on the matching degree between question and answer representations as well as the influence of respondents.  
However, they encounter two main limitations: (1) Correlations between answers in the same question are often overlooked. (2) Question and respondent representations are built independently of specific answers before affecting answer representations.
To address the limitations, we devise a novel graph-based tri-attention network, namely GTAN, which has two innovations. 
First, GTAN proposes to construct a graph for each question and learn answer correlations from each graph through graph neural networks (GNNs).
Second, based on the representations learned from GNNs, an alternating tri-attention method is developed to alternatively build target-aware respondent representations, answer-specific question representations, and context-aware answer representations by attention computation.
GTAN finally integrates the above representations to generate answer ranking scores.
Experiments on three real-world CQA datasets demonstrate GTAN significantly outperforms state-of-the-art answer ranking methods, validating the rationality of the network architecture.
\end{abstract}

\section{Introduction}
\begin{figure}[!t]
    \centering
    \subfloat[Answer Occurrence]
    {\includegraphics[width=0.45\linewidth]{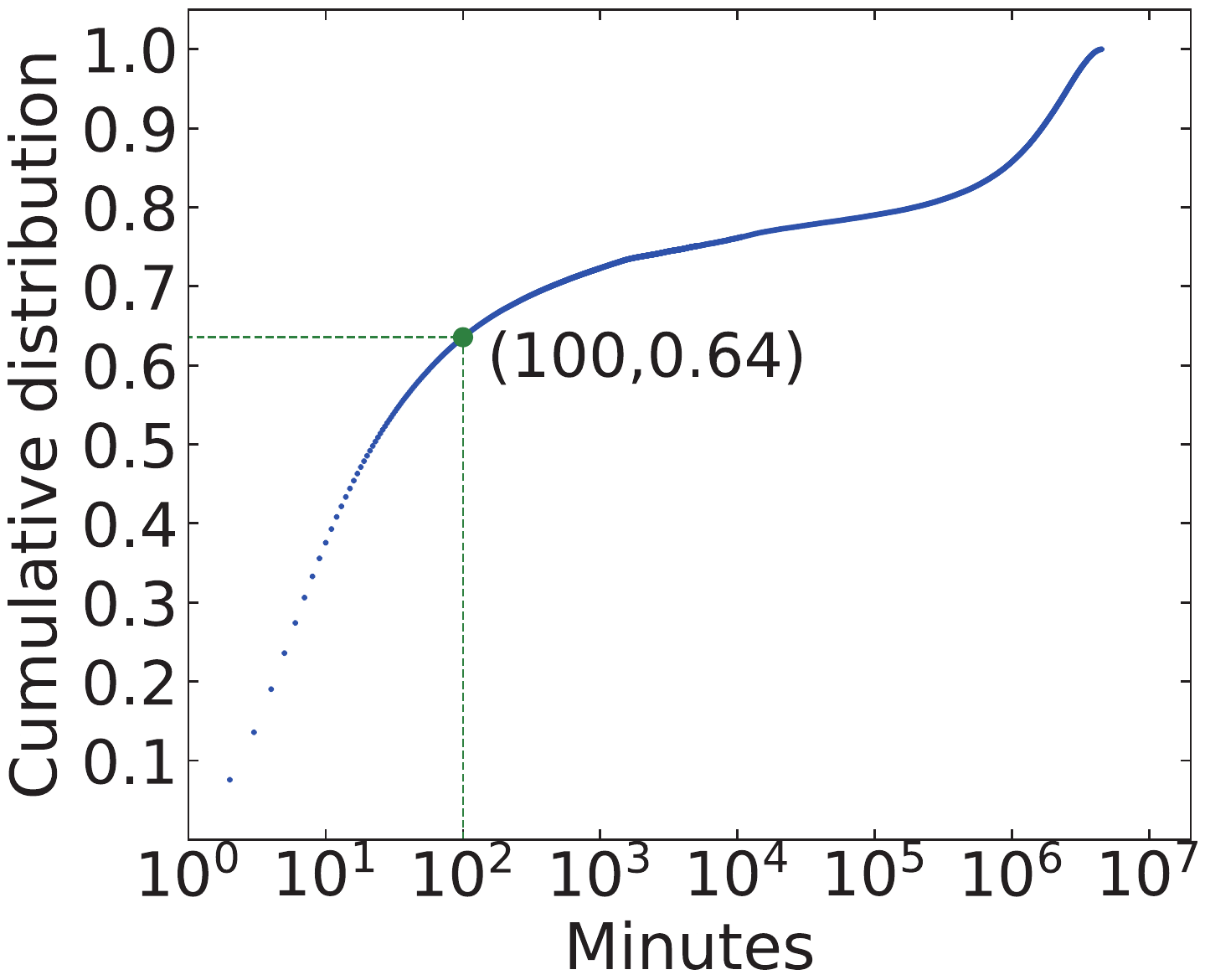}\label{fig:occurrence_so1}}    
    \hspace{1mm}
    \subfloat[Vote Occurrence]
    {\includegraphics[width=0.45\linewidth]{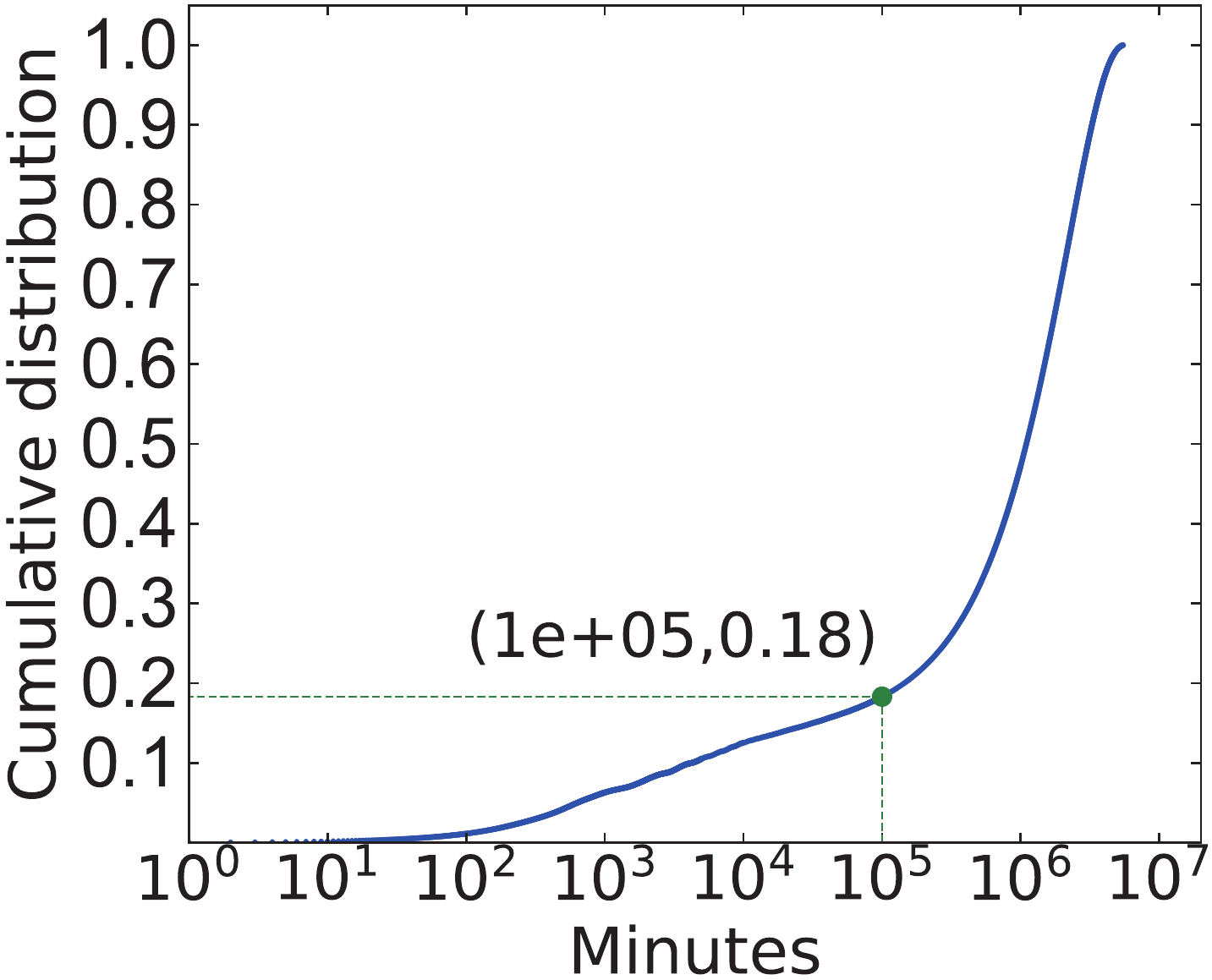}\label{fig:occurrence_so2}}    
    \caption{Distributions of the occurrence of answers and votes with respect to time interval in StackOverflow.}
    \label{fig:data_stats_so}
\end{figure}

Community-based question answering (CQA) is a pivotal online service gathering the wisdom of the crowd.
It enables users to ask and answer questions, and draws much research interest~\citep{ji-cikm2012}.
Popular CQAs, including StackOverflow, Quora, and Zhihu, have been the crucial entries for modern knowledge sharing and retrieval due to the accumulated millions of questions and answers.
Consequently, it is of great importance for CQA platforms to support high-quality answer selection for ordinary users.
Although the votes (e.g., thumbs-up) received by each answer can be the standards to prioritize high-quality answers, their accumulation inevitably needs notable time costs (see Figure~\ref{fig:data_stats_so}).
As such, automatic answer ranking is indispensable for quickly finding potentially popular answers with high quality.

To this end, some conventional approaches~\citep{ShahP10,HuLWLW13,tymoshenko-2015icikm,OmariCRS16} conduct tedious feature engineering to find effective features, whereas they are labor-intensive and have limited generalization ability.
Thus recent efforts~\citep{shen2015question,qiu2015convolutional,fang2016community,zhang2017attentive,zhao2017community,HuQFX18,lyu2019we} are devoted to deep representation learning approaches for measuring the matching degree between target questions and answers.
In particular, a few recent studies~\citep{zhao2017community,HuQFX18,lyu2019we,xie2020attentive} additionally build respondent representations based on their IDs or historical answers, hoping to reveal user expertise on answer quality.
It is worth noting that the respondent role embodies the unique characteristics of CQA, as compared to general question answering.

\begin{figure}[!t]
	\includegraphics[width=\linewidth]{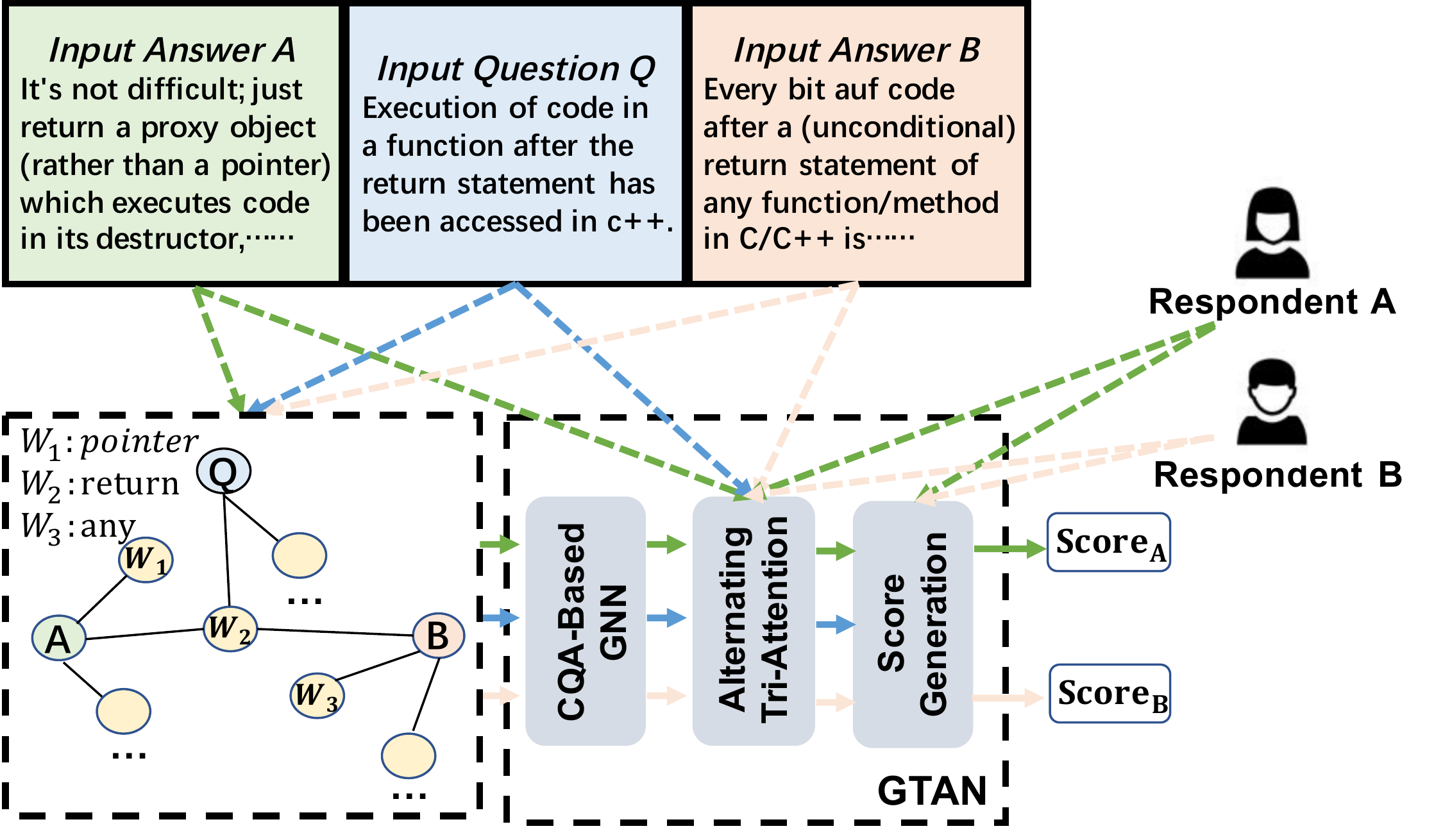}
    \caption{An illustrative example of automatic answer ranking in CQA for one question and its answers.}\label{fig:sketch}
\end{figure}

However, the above approaches are subject to two inherent limitations.
Firstly, they learn each answer representation independently, yet the correlations between answers belonging to the same question are overlooked, which provide some clues in characterizing answers (see Figure~\ref{fig:sim} where top-ranked answers associated with the same question have larger similarities).
Secondly, each question and respondent are assumed to only have one context-free representation for affecting answer representations, which does not conform to real situations.
On the one hand, different parts of a question might be addressed by its different answers.
As shown in Figure~\ref{fig:sketch}, answer A gives more information w.r.t. the word ``execution'' in the question, while answer B concentrates more on the word ``statement''.
Thus it is more promising to build answer-specific question representations to further promote answer representations.
On the other hand, a respondent is usually with mixed professional interests, but only a part of them is associated with a given question-answer pair.
As such, it might be better to build target-aware representation for a respondent. 

To address the above two limitations, we propose the novel GTAN model (short for \underline{G}raph-based \underline{T}ri-\underline{A}ttention \underline{N}etwork) to take questions, answers, and respondents as input, and output answer ranking scores.
As shown in Figure~\ref{fig:sketch}, we first construct a graph for the question and its answers, wherein word nodes act as bridges to connect answers.
Consequently, GTAN customizes a CQA-based graph neural network (GNN) to encode answer correlations into their representations.
Based on the graph-based representations obtained by GNN, an alternating tri-attention method is developed to first build target-aware respondent representations based on QA-guided attention gating and answer-specific question representations through answer-guided question attention.
Then the two types of representations in turn affect answers to form context-aware answer representations by question and respondent-guided answer attention.
As such, representations of questions, respondents, and answers are alternatively updated. 
GTAN ultimately integrates the above-obtained representations to compute answer scores.
In summary, we make the following contributions:

(1) We highlight the two limitations of existing answer ranking methods by showing the necessities of encoding answer correlations into answer representations and learning target-aware respondent representations and answer-specific question representations.

(2) We propose GTAN that contains two innovations: a customized GNN for encoding answer correlations and an alternating tri-attention mechanism to learn question, respondent, and answer representations.
To our knowledge, this is the first study to build heterogeneous graphs and applying GNN for answer ranking in CQA.

(3) We demonstrate GTAN achieves the superior performance through extensive experiments on three real-world datasets and validate overcoming each limitation indeed promotes answer ranking performance.

\section{Related Work}\label{sec:related}
Existing approaches for answer ranking in CQA are mainly categorized into aspects: \textit{feature-based} approaches and \textit{representation learning-based} approaches.
The first category heavily relies on manual-crafted features. 
An early study~\citep{ShahP10} proposes to use classification models (e.g., logistic regression) with the input of the features constructed from questions, answers, and respondents, such as the length of answer text and the number of questions answered by a respondent.
More complex and advanced textual features, such as dependency-based structural features~\citep{tymoshenko-2015icikm} and novelty based features~\citep{OmariCRS16} have also been investigated to improve their performance.
Besides, more than forty Boolean features are exploited to classify two answers into the same class or not in a separate stage~\cite{JotyBMFMMN15}.
However, these feature-based approaches have low generalization ability due to the domain-specific features and are labor-intensive. 

To alleviate the above issues, representation learning-based approaches have become the paradigm.
Most of them regard the problem as question-answer text matching and learn low-dimensional feature representations for questions and answers.
The work~\citep{shen2015question} calculates word-level cosine similarities based on word embeddings of questions and answers obtained by Skip-gram~\citep{MikolovSCCD13}.
Neural tensor network~\citep{SocherCMN13} is further utilized to model the matching degree with a non-linear tensor layer~\citep{qiu2015convolutional}. 
To incorporate the role of respondents, a few recent studies learn their representations so as to better characterize the matching between questions and answers.
The work~\citep{HuQFX18} considers multi-modal content (e.g., text and image) and social relations between respondents to enrich the representations of their corresponding questions and answers.
Both the studies~\citep{zhao2017community} and~\citep{lyu2019we} decompose the matching computation into two parts: question-answer matching (the same as previous studies), and question-respondent matching (newly added to incorporate the effect of respondents).
The enhancement part of the study~\citep{lyu2019we} over the work~\citep{zhao2017community} is attributed to the answer representation learning with latent user expertise and hierarchical attention mechanisms, as compared to the previous work that models answer embeddings independent on their authors.
Moreover, AUANN~\citep{xie2020attentive} employs generative adversarial networks (GANs)~\cite{GoodfellowPMXWOCB14} to help to learn from user past answers for acquiring relevant user representations, which are then directly fed into score computation.

As aforementioned, the above representation learning-based models have two limitations which motivate this study to pursue more effective representations.
GNNs~\citep{kipf2016semi,velivckovic2017graph}, which have already been applied to general question answering~\cite{CaoAT19,TuWHTHZ19}, and attention mechanism~\citep{BahdanauCB14,ZhangWWZ18} are the backbones to devise the CQA-based graph neural network for the first limitation and alternating tri-attention mechanism for the second limitation, respectively.

\section{Preliminaries}\label{sec:pre}
\subsubsection{Problem Formulation}\label{subsec:problem}
Assume we have a question set $\QSet^{tr}$ for training.
Taking question $q\in\QSet^{tr}$ as an illustrative example throughout this paper, it consists of a textual description of length $l(q)$, denoted as $(w^q_{1},w^q_{2},...,w^q_{l(q)})$, and a set of answers, denoted as $\ASet_q=\{a^q_1,a^q_2,...,a^q_{n(q)}\}$, where $n(q)$ is the number of answers.
In reality, each answer is provided by a user (a.k.a. respondent).
Hence we have a corresponding user set $\USet_q=\{u^q_1,u^q_2,...,u^q_{n(q)}\}$ for question $q$.  
Furthermore, we denote $a^q_i$ as a composition of $l(a^q_i)$ words, i.e., $a^q_i=(w^q_{i,1},w^q_{i,2},...,w^q_{i,l(a^q_i)})$. 
All the words come from a pre-defined vocabulary set $\Vocab$.

Given the above notations, the aim of automatic answer ranking in CQA is to learn a function: $f(q, a^q_i, u^q_i) \rightarrow s^q_i$ to generate score $s^q_i$ of answer $a^q_i$, which is provided by user $u^q_i$ for question $q$.
Since the set of ground-truth vote counts, i.e.,  $\VoteSet_q = \{v^q_1,v^q_2,...,v^q_{n(q)}\} (v^q_i \in \ZSet^*)$ is known in the training set, it is used to train the score function.
To be specific, given two answers $a^q_i, a^q_j\in \ASet_q$ satisfying $v^q_i > v^q_j$, we aim to learn a score function $f(\cdot)$ to make the scores $s^q_i$ and $s^q_j$ satisfy the following inequality: $s^q_i > s^q_j$.
The above procedure holds true for every question in $\QSet^{tr}$.
Through the learned score function, we can generate scores to automatically rank answers for a newly arrived question.
Without causing ambiguity, we omit the superscript $q$ in the remaining of the paper for simplicity. 

\subsubsection{Data Analysis}\label{subsec:data-analysis}
In this part, we conduct preliminary data analysis for the following two aspects: 1) the distribution of time intervals w.r.t. questions, answers, and votes; 2) the existence of answer correlations. 
We have three real-world CQA datasets (i.e., StackOverflow, Zhihu, and Quora) for analysis, the details of which will be illustrated in the experimental section. 

\begin{figure}[!t]
    \centering
    \subfloat[StackOverflow]
    {\includegraphics[width=0.32\linewidth]{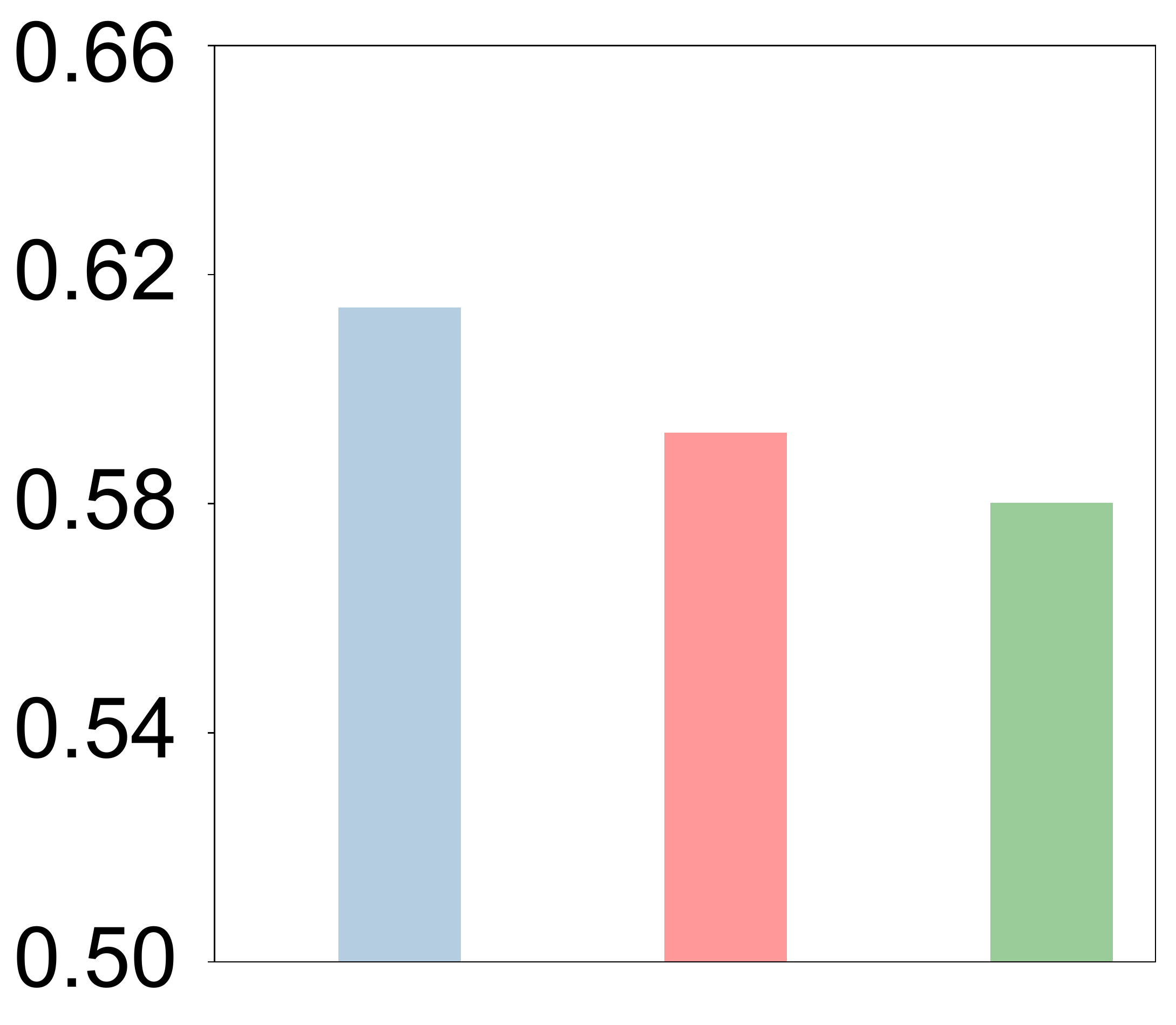}\label{fig:sim_so_3}}
    \subfloat[Zhihu]
    {\includegraphics[width=0.32\linewidth]{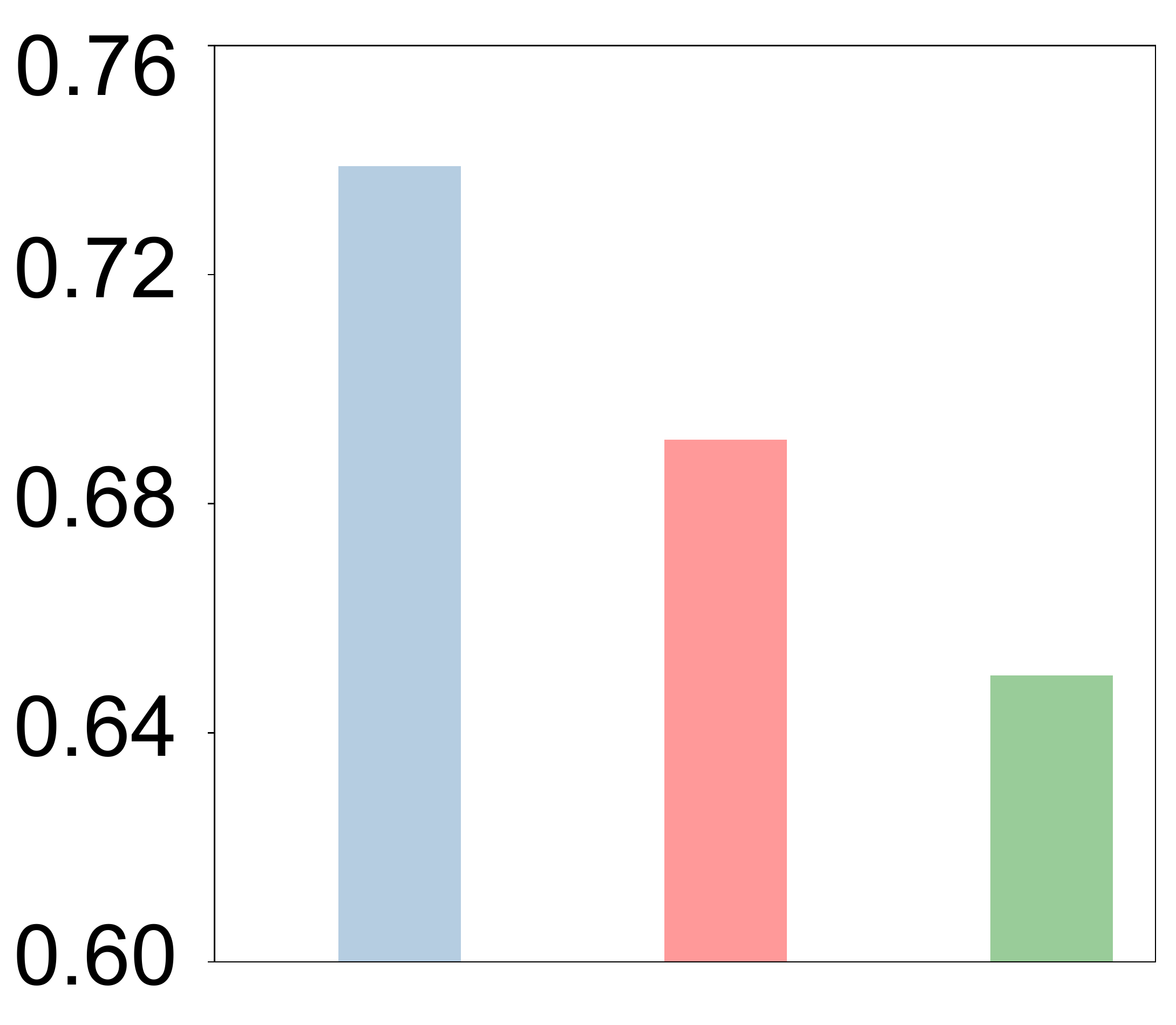}\label{fig:sim_zhihu_3}}
    \subfloat[Quora]
    {\includegraphics[width=0.32\linewidth]{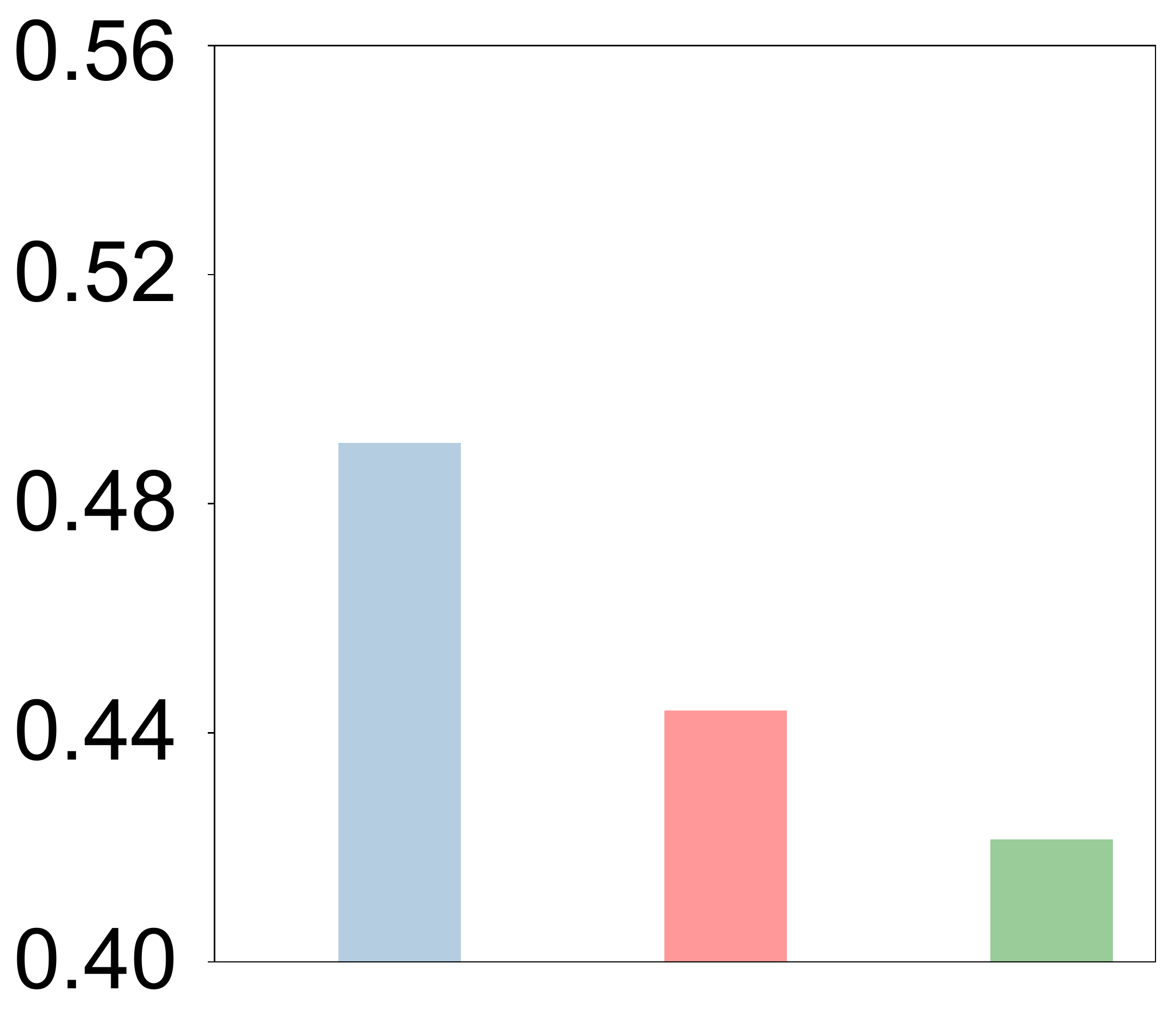}\label{fig:sim_quora_3}}
    \hspace{-10mm}
    \subfloat{\includegraphics[width=0.8\linewidth]{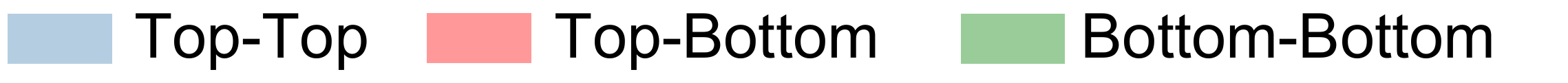}}
    \caption{Textual similarities between answers with similar or different quality levels.}
    \label{fig:sim}
\end{figure}

Due to the lack of vote timestamps in Zhihu and Quora, we only take StackOverflow for illustrating the first aspect.
We first depict the distributions of time intervals between the timestamps when a question is raised and when its answers occur.
As Figure~\ref{fig:occurrence_so1} shows, about 64\% of the answers are posted within $100$ minutes after their questions are raised, accounting for a large proportion of all the answers.
By contrast, in Figure~\ref{fig:occurrence_so2}, merely 18\% of the votes are collected within a time interval of $10^5$ minutes (over $2$ months) after the corresponding answers are provided.
The disproportionate distribution implies that far before we can observe enough votes for answer ranking, an automatic answer ranking mechanism is needed to return potentially popular answers, which lays the foundation of the problem.

Furthermore, the existence of answer correlations is verified by comparing the textual similarities calculated on answers with different types of ranks.
To accomplish this, we represent each answer through a pretrained Doc2Vec model~\citep{le-icml2014doc2vec} and use cosine similarity, which is insensitive to text length.
Given answers' vote number, we rank them in descending order and thus high-quality answers are in top positions. 
We regard the answers in the range of the first 25-th percentile as ``Top'' ones and those after 75-th percentile as ``Bottom'' ones. 
For each question, we compute the average cosine similarities between ``Top'' and ``Top'' answers, ``Top'' and ``'Bottom' answers, and ``Bottom'' and ``Bottom'' answers.
Figure~\ref{fig:sim} reports that the similarities between ``Top'' and ``Top'' answers (i.e., ``Top-Top'') are consistently larger than the other two kinds of similarities in the three datasets.
The law behind this phenomenon might be that high-quality answers belonging to the same question share some spirits, which is reflected by textual similarities.
We also measure the cosine similarity between different types of answer corresponding to the figure.
The similarity values w.r.t. "Top-Top" are greater than those w.r.t. "Bottom-Bottom" by 30.5\%, 52.5\%, and 20.9\% on the StackOverflow, Zhihu, and Quora datsets, respectively.
Therefore the intuition of good answers with larger similarities then inspires us to build graphs to learn answer correlations through representation propagation.

\begin{figure}[!t]
    \centering
	\includegraphics[width=\linewidth]{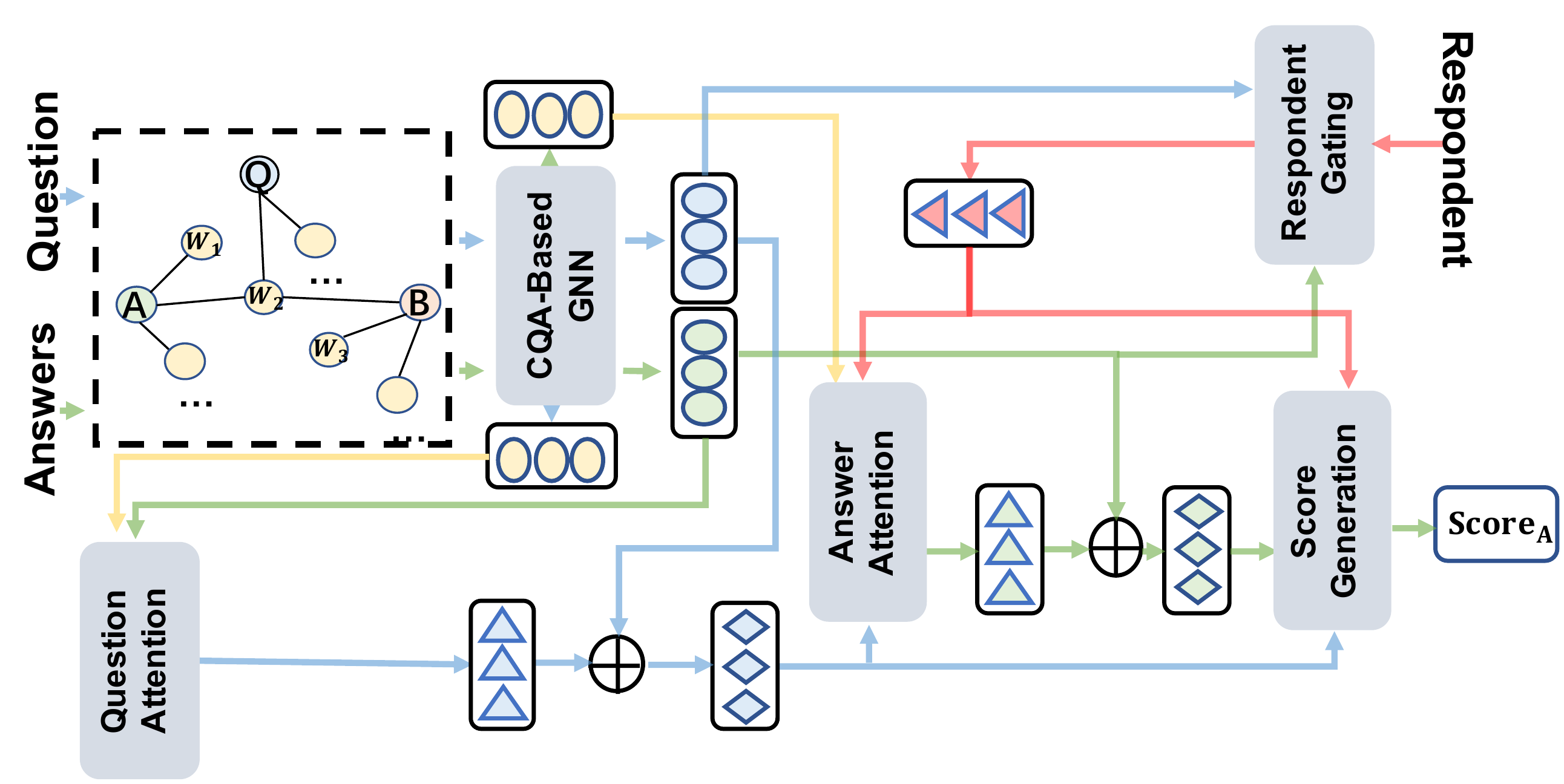}
    \caption{Architecture of GTAN (blue: questions, green: answers, red: respondents, yellow: words).}
    \label{fig:architecture}
\end{figure}

\section{Proposed Methodology: GTAN}

\noindent\textbf{Model Overview:}
The architecture of GTAN is depicted in Figure~\ref{fig:architecture}.
It briefly presents the information flow from input question, answer, and respondent, to answer score.
Basically speaking, the CQA-based graph neural network is first executed for correlation modeling and getting the graph-based question, answer, and word representations.
Consequently, respondent gating and question attention are conducted to gain target-aware respondent representations and answer-specific question representations, which are in turn used for building context-aware answer representations through answer attention.
Based on the above-obtained representations, answer score generation can be ultimately performed. 

\subsection{CQA-Based Graph Neural Network}
As the graph example shown in Figure~\ref{fig:architecture},  we build a text graph for each given question, which has three types of nodes, i.e., question, answer, and word.
It is intuitive that answer correlations are reflected by their shared words, and this is also true for the relevance between a question and its answers.
With this intuition, we have answer-word edges and question-word edges.
The weights are assigned by term frequency-inverse document frequency (TF-IDF) to capture the word importance for answers and questions.
For ease of calculation, the adjacency matrix $\AMAT$ is used to contain the TF-IDF weights of the graph, wherein $\AMAT_{i,j}=1$ if $i=j$.

Before introducing the CQA-based GNN, we first assume the input representations of question $q$, question words $w_j (j\in \{1,...,l(q)\})$, answers $a_i (i\in\{1,...,n(q)\})$, and answer words $w_{i,j} (j\in \{1,...,l(a_i)\})$ for each $a_i$ are represented as $\eVEC$, $\eVEC_{w_j}$, $\eVEC_{a_i}$, and $\eVEC_{w_{i,j}}$, respectively. 
They are fixed during the training stage, so as to remain consistent with testing on new questions and answers.
For ease of illustration, we use an integrated embedding matrix $\EMAT^0$ to include all of them, where $\EMAT^0_i$ indicates the $i$-th node of the graph.

Firstly, taking node $i$ as an example, CQA-based GNN aggregates the representations of its neighboring nodes as:
\begin{equation}\label{eq:aggre}
   \HMAT^0_i =\sum_{j\in N_r(i)}\AMAT_{i,j}\EMAT^0_j\,,
\end{equation}
where $N_r(i)$ denotes the type-$r$ neighboring set of node $i$.

Secondly, CQA-based GNN updates the target node representation by differentiating node types as follows:
\begin{equation}\label{eq:update}
    \bar{\EMAT^1_i}=\ReLU(\WMAT^1_{\tau}[\EMAT^0_i\oplus\HMAT^0_i\oplus[\EMAT^0_i\odot\HMAT^0_i]])\,,
\end{equation}
where $[~\oplus~]$ denotes a row-wise concatenation and $[~\odot~]$ represents the element-wise product.
$\WMAT^1_{\tau}$ is a type-specific transformation matrix for either questions, answers, or words (i.e., $\tau\in\{1,2,3\}$), thus keeping aware of the heterogeneity of the graph. 
Noting that CQA-based GNN is simpler than complex heterogeneous graph neural networks~\citep{ZhangSHSC19} and recent studies~\citep{WuSZFYW19,He2020sigir} demonstrate simplified GNNs could retain comparable performance as compared to complex GNNs.
To relieve over-smoothing issue~\citep{li2019deepgcns}, we adopt a gate to control information flows from previous layer.
\begin{equation}
    \EMAT^1_i = \alpha*\bar{\EMAT^1_i}+(1-\alpha)*\EMAT^0_i\,,
\end{equation}
\begin{equation}
    \alpha = \sigma(\WMAT^G[\bar{\EMAT^1_i}\oplus\EMAT^0_i]+\bVEC^G)\,,
\end{equation}
where $\WMAT^G$ and $\bVEC^G$ are trainable parameters, and $\sigma$ stands for sigmoid function.

The above two procedures are repeated multiple times to realize high-order propagation. 
Supposing the number of propagation is set to $T$, GTAN ultimately returns the embedding matrix $\EMAT^T$, from which we extract the \textbf{graph-based representations} $\widehat{\eVEC}$, $\widehat{\eVEC}_{a_i}$, $\widehat{\eVEC}_{w_j}$, and $\widehat{\eVEC}_{w_{i,j}}$ corresponding to the input representations.
The trainable parameters in the module are summarized as $\Theta^G = \{\WMAT^G, \bVEC^G, \WMAT^t_{1},\WMAT^t_{2},\WMAT^t_{3}\}|_{t=1}^T$.

\subsection{Alternating Tri-Attention Mechanism}
This mechanism contains QA-guided attention gating for respondent representations, answer-guided question attention for question representations, and question and respondent-guided answer attention for answer representations.

\subsubsection{QA-guided Attention Gating}\label{sec:res-gate}
To capture the expertise relevant to the target question and answer, QA-guided attention gating is proposed to filter irrelevant information in each dimension of the respondent representation.
Specifically, given the ID of respondent $u_i$, a look-up table manner is conducted over respondent embedding matrix $\EMAT^U$ to get original respondent representation $\eVEC_{u_i}$.
Note that we focus on respondents that already have answers in this paper and regard modeling new respondents as future work.
Consequently, the attention gating calculates \textbf{target-aware respondent representation} $\bar{\eVEC}_{u_i}$ based on graph-based question and answer representations, which is defined as follows: 
\begin{equation}
    \bar{\eVEC}_{u_i} = \sigma(\WMAT^U[\widehat{\eVEC}\oplus\widehat{\eVEC}_{a_i}]+\bVEC^U)\odot\eVEC_{u_i}
\end{equation}
where $\Theta^U=\{\WMAT^U,\bVEC^U,\EMAT^U\}$ are trainable parameters.

\subsubsection{Answer-guided Question Attention}\label{sec:que-att}
Using a unique vector for each question limits its expressive ability when affecting the representations of its different answers. 
Thus we propose to compute \textbf{answer-specific question representation} $\widetilde{\eVEC}_{i}$. 
It is realized by regarding graph-based answer representations as queries in the following attention computation.
\begin{equation}\label{eq:que-att}
    \alpha_j=\Softmax(\omega^Q \tanh(\WMAT^Q[\widehat{\eVEC}_{a_i}\oplus\widehat{\eVEC}_{w_j}]+\bVEC^Q)),
\end{equation}
\begin{equation}\label{eq:que-rep}
    \widetilde{\eVEC}_{i}=\sum_{j=1}^{l(q)} \alpha_j \widehat{\eVEC}_{w_{j}},
\end{equation}
where $\alpha_j$ is an attention weight.
$\Theta^Q=\{\omega^Q, \WMAT^Q, \bVEC^Q\}$ are trainable parameters.
Eq.~\ref{eq:que-att} indicates that if a question word has a closer relationship with the answer, it will contribute more to the question representation. 
Though quite simple, answer-specific question representations are indeed beneficial (see Table \ref{tbl:ablation_so}).
In local tests, we have tried to incorporate respondent representations into queries but gain no significant improvements.
This is intuitive since the major role of respondents is the expertise mainly reflected in the answer part. 
At last, we integrate answer-specific question representation $\widetilde{\eVEC}_{i}$ with graph-based question representation $\widehat{\eVEC}$ for answer $a_i$, i.e., $\bar{\eVEC}_{i}=[\widetilde{\eVEC}_{i}\oplus\widehat{\eVEC}]$.

\subsubsection{Question and Respondent-Guided Answer Attention}
We provide answer attention here to grasp the impact of the respondent and question on the answer representation.
It takes $\bar{\eVEC}_{i}$ and $\bar{\eVEC}_{u_i}$ together as the query.
As a result, the computation of the attention weight $\beta_j$ over the target answer word representation $\widehat{\eVEC}_{w_{i,j}}$ is formulated as follows:
\begin{equation}\label{eq:ans-att}
    \beta_j=\Softmax(\omega^A \tanh(\WMAT^A[\bar{\eVEC}_{i}\oplus\bar{\eVEC}_{u_i}\oplus\widehat{\eVEC}_{w_{i,j}}]+\bVEC^A)).
\end{equation}
It reveals that an important answer word should have greater relevance to both the question and the respondent.  
Similarly, we have $\Theta^A=\{\omega^A, \WMAT^A, \bVEC^A\}$ as learnable parameters. 
The \textbf{context-aware answer representation} $\widetilde{\eVEC}_{a_i}$ is gotten by:
\begin{equation}\label{eq:ans-rep}
    \widetilde{\eVEC}_{a_i}=\sum_{j=1}^{l(a_i)} \beta_j \widehat{\eVEC}_{w_{i,j}}.
\end{equation}
Although the methodological aspects in alternating tri-attention mechanism shares partially similar spirits with existing studies attention-based studies~\cite{LuYBP16,NamHK17,ZhangWWZ18}, it is seamlessly integrated with the GNN part for utilizing the graph-based representations to alternately learn effective respondent, question, and answer representations, showing some novel insights.
Finally, we obtain an integrated answer representation
$\bar{\eVEC}_{a_i}=[\widetilde{\eVEC}_{a_i}\oplus\widehat{\eVEC}_{a_i}]$.

\subsection{Score Generation and Training}

\subsubsection{Answer Score Generation}
To generate the ranking score of answer $a_i$, we first concatenate the three representations $[\bar{\eVEC}_{i}\oplus \bar{\eVEC}_{a_i} \oplus \bar{\eVEC}_{u_i}]$ to have $\zVEC_{i}$.
Afterwards, we feed $\zVEC_{i}$ into a feed-forward neural network with $K$ fully-connected (FC) layers.
Hence the corresponding score is calculated as:
\begin{equation}\label{eq:score}
    s_i = \FC_K\Big(\big(\cdots\FC_1(\zVEC_{i};\theta_1);\cdots\big); \theta_K\Big),
\end{equation}
where $\FC_k$ denotes the $k$-th fully-connected layer.
As usual, we summarize the trainable parameters as $\Theta^F = \{\theta_1,...,\theta_K\}$.
In summary, we have the score function $f(q,a_i,u_i;\Theta)$ to generate answer ranking score $s_i$, where $\Theta=\{\Theta^G, \Theta^U, \Theta^Q, \Theta^A, \Theta^F\}$ cover all trainable model parameters.

\subsubsection{Model Training}
The aim of model training is to learn optimal parameters $\Theta$ by referring to the ground-truth vote counts of answers.
By convention~\citep{zhao2017community,lyu2019we}, we adopt pairwise learning to rank and define the loss function for question $q$ ($q\in\QSet^{tr}$) as follows:
\begin{equation}\label{eq:loss}
    \Loss = \sum_{i,j~(v_i>v_j)} \max\Big(0, c+f(q,a_j,u_j;\Theta)-f(q,a_i,u_i;\Theta)\Big),
\end{equation}
where $c$ ($c=1$ for experiments) is the specified margin for pairwise ranking.
The initialization of the input representations (e.g., word embeddings) and the used optimization algorithm, as well as some hyper-parameter settings are clarified in the experimental part.

\section{Experiments}
In this section, we elaborate the experimental setup and analyze the experimental results, aiming to answer:

\noindent \textbf{\texttt{RQ1}}: Can GTAN achieve better answer ranking performance than the state-of-the-art methods for answer ranking?
	
\noindent\textbf{\texttt{RQ2}}: How do the key model components and information types used in GTAN contribute to the overall performance?

\subsection{Experimental Setup}\label{sec:exp-setup}

\subsubsection{Datasets}
To ensure the reliability of the results, we conduct experiments on three real-world datasets, i.e., StackOverflow\footnote{\url{https://archive.org/download/stackexchange/stackoverflow.com-Posts.7z}}, Zhihu (collected from its website), and Quora~\citep{lyu2019we}.
They correspond to three representative and complementary CQA platforms.

For all the Chinese text in Zhihu, we adopt Jieba\footnote{\url{https://github.com/fxsjy/jieba}} for word segmentation.
To filter some noisy data, we adopt the following procedures: (1) We filter respondents with less than 5 answers.
(2) We remove answers with less than 5 words and questions with less than 5 answers or too many answers (e.g., 1000).
(3) We discard words that appear infrequently (e.g., less than 10 times).
The above procedures could be repeated several times to have a stable data size. 
Finally, the basic statistics of the three datasets are summarized in Table~\ref{tbl:data_stats}.
Particularly, the constructed graphs have on average over 200 nodes for the StackOverflow and Quora datasets and over 1000 nodes for the Zhihu dataset.
We split the datasets into training sets, validation sets, and test sets, according to the ratios of about 8 to 1 to 1.

\begin{table}[!t]
	\centering
\resizebox{\linewidth}{!}{
	\begin{tabular}{c|ccccc}
		\toprule[0.8pt]
		Dataset     &\#Que. &\#Ans. &\#Resp. &Vocab. & Avg. Len. \\
		\midrule[0.6pt]
		SO &139128    &884261   &40213   &52457   &81.6 \\
		Zhihu         &19357     &473338   &133203  &65880   &85.6 \\
		Quora        &12664     &66779    &6061
		&44149  &64.4 \\
		\bottomrule[0.8pt]
	\end{tabular}
	}
	\caption{ Detailed statistics of the three datasets. Avg. Len. denotes the average length of answers.} 
	\label{tbl:data_stats}
\end{table}

\begin{table*}[!ht]
\centering
\begin{tabular}{c|ccc|ccc|ccc} 
\toprule[0.8pt]
\multirow{2}*{\textbf{Method}}& \multicolumn{3}{c}{StackOverflow}& \multicolumn{3}{c}{Zhihu}&\multicolumn{3}{c}{Quora}\\\cline{2-10}
&$\Precision@1$ &MRR &NDCG@3 &$\Precision@1$ &MRR &NDCG@3 &$\Precision@1$ &MRR &NDCG@3\\\midrule[0.6pt]

Doc2vec
& 0.2985 & 0.5136  & 0.6664 & 0.1658 & 0.3555 &  0.3987 & 0.4387  & 0.6336 & 0.7179 \\

S-matrix
& 0.3123 & 0.5353  & 0.6751 & 0.1817  & 0.3653 & 0.4022 & 0.4393  & 0.6383 & 0.7234 \\

CNTN
& 0.3142 & 0.5400  & 0.6823 & 0.2013  & 0.3849 & 0.4275 & 0.4431  & 0.6414 & 0.7278\\

TextGCN
& 0.3248 & 0.5475  & 0.6867 & 0.2011  & 0.3739 & 0.4192 & 0.4128 & 0.6174 & 0.7032 \\  

AMRNL
& 0.3884 & 0.5971  & 0.7327 & 0.3134  & 0.4937 & 0.5657 & 0.6856  & 0.8110 & 0.8811 \\

UEAN
& 0.3967 & 0.6068  & 0.7439 & 0.3354  & 0.5130 & 0.5887 & 0.6877  & 0.8123 & 0.8828 \\

AUANN
& 0.4066 & 0.6132 & 0.7473 & 0.3418 & 0.5188 & 0.5958 & 0.6933 & 0.8152 & 0.8889 \\  \hline

\textbf{Ours (GTAN)} & \textbf{0.4230} & \textbf{0.6265}  & \textbf{0.7597}  & \textbf{0.3556}  & \textbf{0.5313} & \textbf{0.6134} & \textbf{0.7235}  & \textbf{0.8368} & \textbf{0.9003}\\

\bottomrule[0.8pt]
\end{tabular}
\caption{Performance comparison of all adopted approaches on the three datasets.}\label{tab:all_result}
\end{table*}

\subsubsection{Evaluation Metrics}
To keep consistent with previous studies~\citep{lyu2019we}, we adopt the following ranking metrics: (1) Mean Reciprocal Rank (MRR), which measures how the best answer is ranked by different approaches; (2) Normalized Discounted Cumulative Gain (NDCG), which provides a position-aware performance; (3) $\Precision@1$ (short for Precision@1), which calculates the ratio that the best answers are ranked at the first position. For each question, the best answer used to calculate MRR and $\Precision@1$ is the one with the largest vote count within that question.
Similarly, the ground-truth positions of answers are determined by sorting their vote counts in descending order.

\subsubsection{Baselines} 
The representative baselines are as follows:

\noindent-\;\textbf{Doc2Vec}~\citep{le-icml2014doc2vec} naturally extends Word2Vec~\citep{MikolovSCCD13} by associating sentence-level representations accompanied by word embeddings.
We utilize Doc2Vec to gain answer representations and add FC layers on top of them to generate answer scores.

\noindent-\;\textbf{S-matrix}~\citep{shen2015question} gains a semantic similarity matrix between a question and an answer in the word level.
It is followed by convolutional layers to model the matrix.

\noindent-\;\textbf{CNTN}~\citep{qiu2015convolutional} characterizes the complex interactions between questions and answers with the combination of convolutional networks and neural tensor networks.
Dynamic k-max pooling~\citep{KalchbrennerGB14} is leveraged within it.

\noindent-\;\textbf{AMRNL}~\citep{zhao2017community} decomposes the similarity computation into the question-answer part and question-respondent part.
Thus the role of respondents is first considered for the studied problem.

\noindent-\;\textbf{UEAN}~\citep{lyu2019we} performs both word-level and sentence-level attention computations, which enable question topic and user expertise to affect answer representations. 
The decomposition of similarity computation is inherited from AMRNL for score generation.

\noindent-\;\textbf{TextGCN} slightly modifies original TextGCN~\citep{yao2019graph}, a heterogeneous graph convolutional network for document classification, to make it applicable to the problem.
Specifically, it replaces classification loss with the pairwise ranking loss shown in Eq.~\ref{eq:loss}.

\noindent-\;\textbf{AUANN}~\cite{xie2020attentive} utilizes GANs to help learning representations of respondents based on their historical answers.
The relationships between questions and answers are considered as well.

All the above baselines are tuned on validation datasets to select the hyper-parameter configurations.

\subsubsection{Implementation of GTAN}\label{sec:exp-impl}
The word representations for CQA-based GNN are initialized by Word2Vec.
We have also tried BERT~\citep{DevlinCLT19} to provide pre-trained embeddings, but no significant gains are observed in the local tests. 
The initial question and answer representations are obtained by performing mean-pooling over their word embeddings, respectively.
Based on the performance of GTAN on validation datasets,
we set the number of propagation layers $T=2$.
The dimension of representations for words, questions, answers, and respondents are all set to 64.
The number of FC layers $K$ is set to 2 for a non-linear transformation.
Adam~\citep{kingma2014adam} is adopted for model optimization, with the initial learning rate of 0.0005.
We implement the models by Tensorflow and run the experiments on a GPU (Nvidia GeForce GTX 1080 Ti) with 11GB memory.
All the results are averaged over three runs.

\begin{table*}[!t]
\centering
\begin{tabular}{l|cc|cc|cc} 
\toprule[0.8pt]
\multicolumn{1}{c|}{\multirow{2}*{\textbf{Method}}}& \multicolumn{2}{c|}{StackOverflow}& \multicolumn{2}{c|}{Zhihu}&\multicolumn{2}{c}{Quora}\\\cline{2-7}
&$\Precision@1$ &MRR &$\Precision@1$ &MRR &$\Precision@1$ &MRR \\\midrule[0.6pt]
\textbf{Ours (GTAN)} & \textbf{0.4230} & \textbf{0.6265}  & \textbf{0.3556}  & \textbf{0.5313}  & \textbf{0.7235} & \textbf{0.8368}\\

~~-~~w/o Graph
& 0.3828 & 0.5958  &  0.3302 & 0.4894 & 0.7037 & 0.8173  \\

~~-~~w/o T-MAT 
& 0.4104 & 0.6156  & 0.3325 & 0.5109  & 0.7117 & 0.8275 \\

~~-~~w/o Que
& 0.4108 & 0.6160  & 0.3372 & 0.5154  & 0.7123 & 0.8289 \\

~~-~~w/o Res
& 0.3211 & 0.5448  & 0.1957 & 0.3784  & 0.4343 & 0.6342  \\

~~-~~w/o Tri-Att
& 0.4027 & 0.6117  & 0.3356 & 0.5029  & 0.7085 & 0.8259 \\

~~-~~w/o Que-Att
& 0.4127 & 0.6166  & 0.3432 & 0.5191  & 0.7162 & 0.8299  \\

~~-~~w/o Res-Att 
& 0.4122 & 0.6171  & 0.3363 & 0.5135  & 0.7130 & 0.8290  \\

~~-~~w/o Res-Gate 
& 0.4109 & 0.6160  & 0.3470 & 0.5202  & 0.7099 & 0.8273  \\

\bottomrule[0.8pt]
\end{tabular}
\caption{Ablation study of GTAN.}\label{tbl:ablation_so}
\end{table*}

\subsection{Experimental Results}

\subsubsection{Overall Comparison (\textbf{\texttt{RQ1}})}
Table~\ref{tab:all_result} shows the overall comparison of GTAN with different baselines.
We observe Doc2Vec performs poorly on the three datasets.
This conforms to expectation since it only considers the answer aspect by learning answer representations, while neither questions nor respondents are utilized.
By further comparing Doc2Vec with S-matrix, we find the performance is improved to a certain degree.
This demonstrates the quality of answer text largely determines the ranking positions and considering the relations between answers and questions could bring additional gains.
Compared to S-matrix, CNTN and TextGCN achieve better performance in most cases.
It makes sense because the two models have more advanced text relevance modeling techniques (i.e., CNTN with neural tensor networks and TextGCN with GCN) than S-matrix which is mainly based on word-level similarities. 

For the models of AMRNL and UEAN, they calculate the relevance between questions and respondents, and achieve significantly better results than the above-mentioned methods which totally overlook the role of respondents.  
This can be attributed to the fact that the expertise of respondents heavily affects the quality of answers regarding specific topics, since respondents might be knowledgeable of different topics. 
Therefore, considering two different modalities of answers, i.e., answer text and respondent, is critical for better performance.
Moreover, UEAN performs better than AMRNL since the former one could acquire more expressive answer representations. 
This is because UEAN incorporates respondent representations into the attention computation over answer words, but AMRNL does not directly correlate answer and respondent representations
AUANN further improves the performance, thanks to the introduction of GANs to learn better respondent representations.

In the end, our model GTAN consistently yields the best answer ranking performance.
To be specific, GTAN significantly improves the best baseline AUANN from $40.66\%$ to $42.30\%$ by $\Precision@1$ on StackOverflow, from $59.58\%$ to $61.34\%$ by $\NDCG@3$ on Zhihu, and from $81.52\%$ to $83.68\%$ by MRR on Quora, which is verified by a paired t-test.
Compared to AUANN, GTAN is capable of explicitly encoding answer correlations into their representations by GNN modeling, and learning answer-specific question representations and target-aware respondent representations, which are promising as demonstrated.

\subsubsection{Ablation Study (\textbf{\texttt{RQ2}})}
To investigate the contributions of key components and information types adopted by GTAN, we make the following variants of GTAN:
(1) \textbf{``w/o Graph''} denotes removing CQA-based GNN, equivalent to feeding input representations directly into alternating tri-attention computation.
(2) \textbf{``w/o T-MAT''} means only using a shared matrix in Eq.~\ref{eq:update}, without differentiating node types.
(3) \textbf{``w/o Que''} represents discarding all question information in GTAN.
(4) \textbf{``w/o Res''} denotes discarding all respondent information.
(5) \textbf{``w/o Tri-Att''} keeps the GNN part but removes tri-attention computation, where the output representations of GNN, along with respondent representations, directly form the integrated representations, as shown in Eq.~\ref{eq:score}.
(6) \textbf{``w/o Que-Att''} only removes the question attention computation, denoting each question has only one unique representation for similarity calculation.
(7) \textbf{``w/o Res-Att''} removes respondent representations from Eq.~\ref{eq:ans-att}, meaning not considering the role of respondents in answer attention computation.
(8) \textbf{``w/o Res-Gate''} removes QA-guided Attention gating, meaning directly using original respondent representations in GTAN.

Table~\ref{tbl:ablation_so} reports the results of the ablation study, from which we have the following key observations:

\noindent * ``w/o Graph'' suffers from noticeable performance degradation, showing the significant contribution of modeling answer correlations for ranking answers.
Therefore we conclude that addressing the \textbf{\textit{first limitation}} is indeed beneficial. 
Moreover, ``w/o T-MAT'' verifies that utilizing type-specific transformation matrices promotes the performance.

\noindent * Both ``w/o Que'' and ``w/o Res'' consistently underperform GTAN, conforming to the fact that in addition to answer text, considering question text and respondents is indispensable. 
One important observation is that ``w/o Res'' performs the worst among the variants, indicating that the expertise of respondents impacts a lot on answer popularity. 

\noindent * The comparison between ``w/o Que-Att'' and GTAN validates the necessity of introducing answer-specific question representations.
By further comparing ``w/o Res-Gate'' with GTAN, we can see obtaining target-aware respondent representations is effective as well. 
As such, we conclude that handling the \textbf{\textit{second limitation}} is effective.

\noindent * In addition, the results of ``w/o Tri-Att'' show the advantages of the involved attention computations as a whole.
And the comparison between ``w/o Res-Att'' and GTAN indicates the effectiveness of considering respondent effect on answer representations (e.g., topic authority).

\begin{figure}[!h]
    \centering
    \subfloat
    {\includegraphics[width=0.32\linewidth]{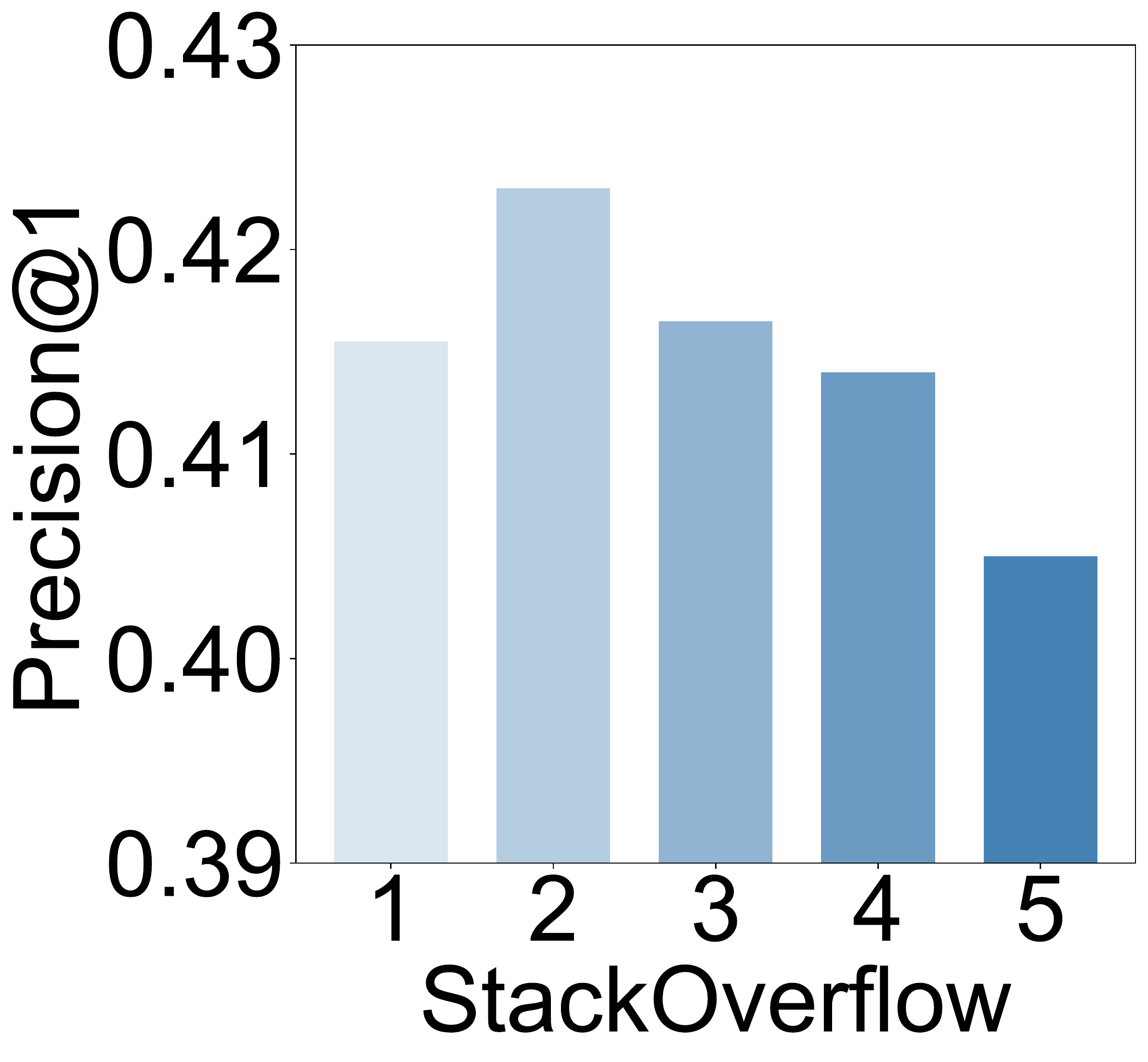}\label{fig:pre1_so_layer}
    \includegraphics[width=0.32\linewidth]{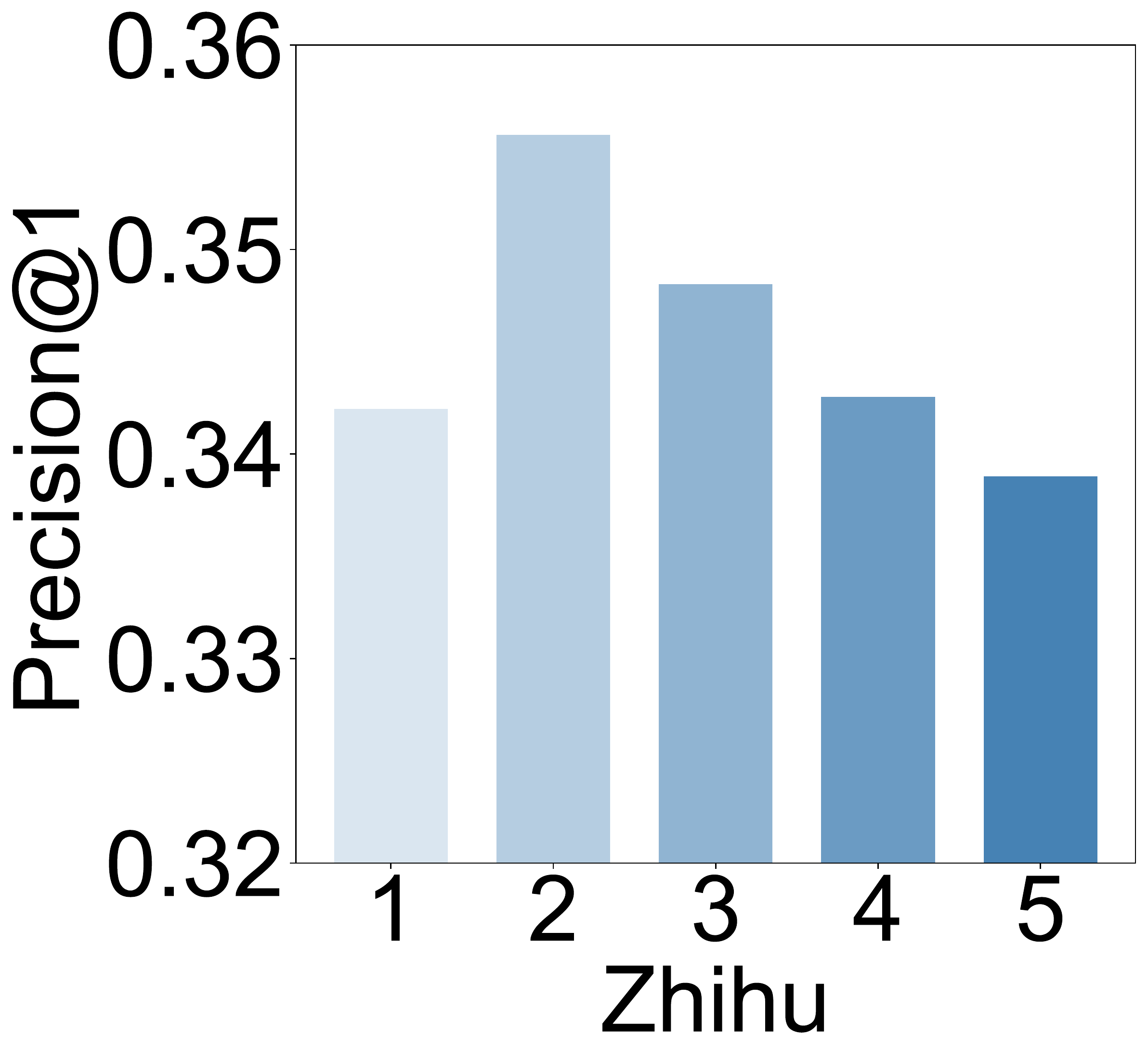}\label{fig:pre1_zh_layer}
    \includegraphics[width=0.32\linewidth]{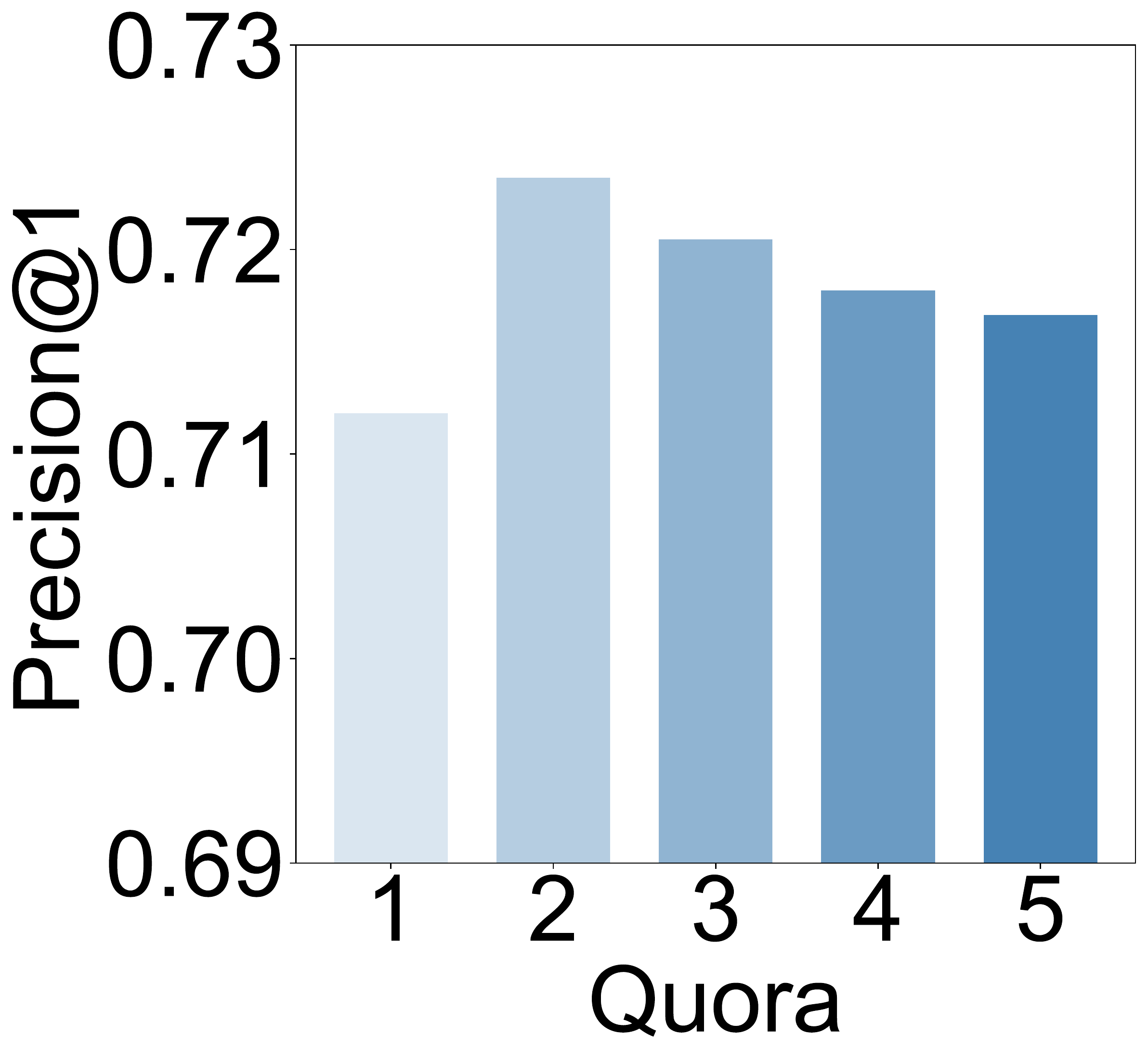}\label{fig:pre1_quora_layer}
    }
    \caption{Result variation w.r.t. different layer number.}
    \label{fig:layer_num}
\end{figure}

\subsubsection{Impact of Propagation Layer Number}
We investigate whether it is beneficial for encoding high-order relations, not just the direct connections between answers and words.  
We test the layer numbers in the range of $\{1,2, 3,4,5\}$.
Figure~\ref{fig:layer_num} depicts the variation trends of the results on the three datasets, from which we observe that:

\noindent * When increasing the layer number from 1 to 2, the improvements are noticeable.
This makes sense because two-layered propagation could enable correlation modeling between different answers via answer-word edges, while one-layered propagation cannot achieve this.
Thus it again verifies the importance of encoding answer correlations.

\noindent * When further adding propagation layers, the performance presents downward trends. The reason might be that adding more layers leads to over-fitting issue and makes answer representations not so distinguishable for judgment.

\subsubsection{Training Efficiency}
We compare the training efficiency of GTAN with the best baseline model AUANN.
Table~\ref{tbl:time} demonstrates that training GTAN gains much higher efficiency than training AUANN.
This phenomenon is reasonable because AUANN demands recurrent neural networks for modeling word sequences, which incurs poor parallel computing in GPUs. 

\begin{table}[!h]
\centering
\begin{tabular}{c|ccc} 
\toprule[0.8pt]
\textbf{Method}&StackOverflow &Zhihu &Quora \\\midrule[0.6pt]

AUANN
& 63.33 & 128.56  &  41.22 \\

\hline

\textbf{Ours (GTAN)} & \textbf{17.02} & \textbf{25.71}  & \textbf{12.95}  \\

\bottomrule[0.8pt]
\end{tabular}
\caption{Average training time cost per question (ms).}\label{tbl:time}
\end{table}

\begin{figure}[!t]
    \centering
    \subfloat[One question and its two answers from StackOverflow.]{
	\includegraphics[width=.95\linewidth]{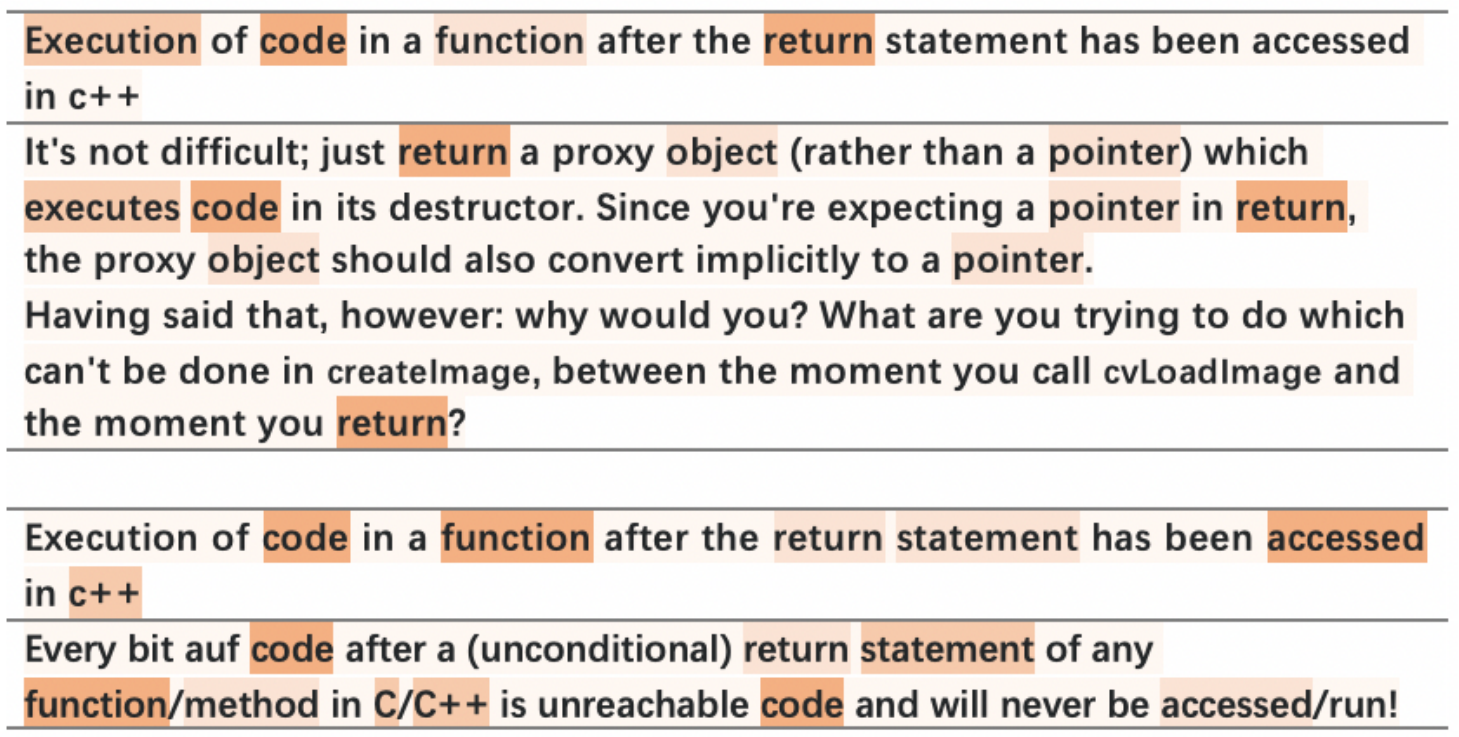}}
	
	\subfloat[One question and its two answers from Zhihu.]{
	\includegraphics[width=.95\linewidth]{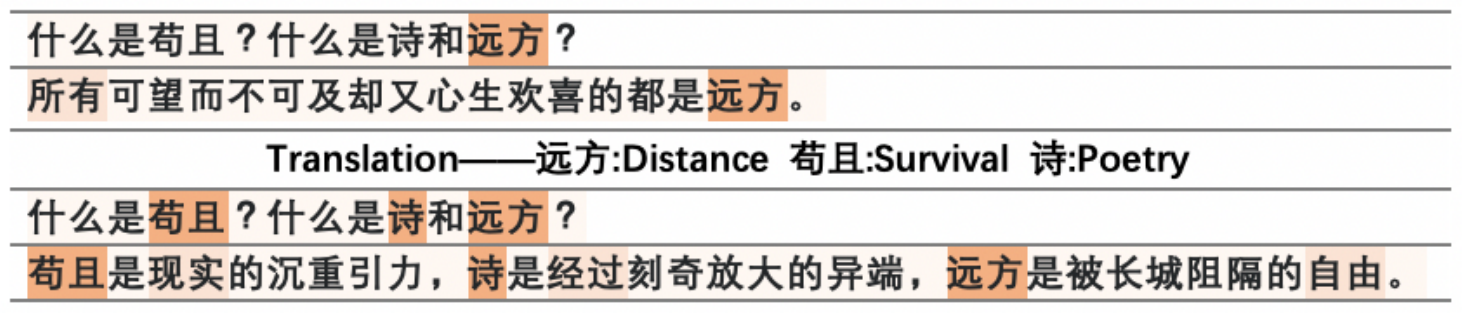}
	}
    \caption{Attention visualization on questions and their answers. Darker colors mean larger attention weights.}
    \label{fig:att}
\end{figure}

\subsubsection{Case Study}
To investigate how the tri-attention mechanism works on real cases, we present one question example and its two answers from StackOverflow and Zhihu, and visualize the attention weights in Figure~\ref{fig:att}.
We observe:
(1) Given a question, the mechanism enables its attention weights to be adaptive to different answers.
As the second question example shows, its question is composed of two parts.
While the first answer only addresses the second part about what is ``Distance'', the second answer presents a more comprehensive view of ``Survival'' in the first part and ``Poetry'' and ``Distance'' in the second part.
As a result, the question representations should be adaptable to their emphasis.
(2) The attention mechanism could find some important words in answers.
For the first example about coding, the two answers have opposite conclusions.
The first answer gives more details about the solutions.
And its contained technical terms ``object'' and ``pointer'' that do not appear in the question text could be addressed by our mechanism.

\section{Conclusion}
In this paper, we study automatic answer ranking in CQA.
We address the two limitations overlooked by the conventional studies by devising a novel graph-based tri-attention network (GTAN).
It has the innovations of combining CQA-based GNN and alternating tri-attention mechanism to learn answer correlations for the first limitation and answer-specific question representations and target-aware respondent representations for the second limitation.
Moreover, GTAN effectively integrates question, answer, and respondent representations for answer score computation.
We conduct extensive experiments on three real-world datasets collected from representative CQA platforms and the results demonstrate the superiority of the proposed model, validating the contributions of the key model components.

\section{Acknowledgments}
The corresponding author Wei Zhang would like to thank the anonymous reviewers for their valuable suggestions.
This work was supported in part by National Natural Science Foundation of China under Grant No. 62072182, 61532010, 61521002, and U1609220, in part by the foundation of Key Laboratory of Artificial Intelligence, Ministry of Education, P.R. China, and in part by Beijing Academy of Artificial Intelligence (BAAI).

\bibliography{reference} 

\end{document}